# Maximally Informative Observables

# and

# Categorical Perception

*An Information-theoretic Formulation*

**Elaine Tsiang**[*]

### Abstract

We formulate the problem of perception in the framework of information theory, and prove that categorical perception is equivalent to the existence of an observable that has the maximum possible information on the target of perception. We call such an observable maximally informative. Regardless whether categorical perception is "real", maximally informative observables can form the basis of a theory of perception. We conclude with the implications of such a theory for the problem of speech perception.

[*]    Monowave Corporation, Seattle, WA., USA

## *Categorical Perception*

The term was coined by Liberman *et al*[Liberman1] to highlight the discovery that an observable of certain speech gestures producing plosive consonants, that of the change in frequency of the second formant, although a continuous variable, leads to the acoustic signals being perceived as one of the three categories of plosive consonants employed in their experiment. The identification function, namely the distribution of the plosive consonants over the variable formant frequency change was close to three indicator functions, one for each plosive (Figure 1[1]).

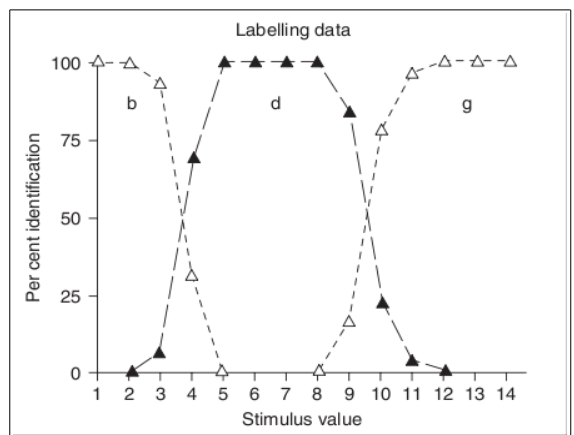

*Figure 1: Identification function for plosive consonants [Goldstone1]*

This was startling because it seems to imply that our senses, with which we are supposed to assess the real world, are heavily influenced, or even determined, by our mental constructs.

Categorical perception has since been observed in other sense modalities in humans, and also in animals. On the other hand, the degree to which the perception is unconditionally categorical has been disputed[Schouten1]. Early explanations attribute the indicator-function-like distribution to the unique psychology of speech. This attribution is unsatisfactory in view of the pervasive and uncertain findings.

Early work relied on subjects' report of their responses, using synthesized speech with relatively coarse sampling resolution of the random variable at issue. Current work[Chang1] measures neural responses directly with cortical surface arrays, with higher sampling resolution and more complex stimuli. Despite all the controversy, categorical perception is undeniably part of how we perceive the world.

We will argue in the following that categorical perception is, far from being hallucinating reality, an efficacious, in fact, the best possible way to observe the real world[2].

---

1  Original diagram by Lieberman, reprinted in [Goldstone1].
2  A more precise statement would be: regardless whether what we call "reality" is itself hallucination, within the limits of our considerations, categorical perception is not an independent, *ad hoc* way of hallucinating.

### *Observations vs. Measurements*

We assume the conventional meaning of a measurement, but distinguish it from an observation in that a measurement directly, or physically, assesses some variable, but an observation assesses a variable as indicative of another variable. So by definition, an observation involves at least two variables, one directly assessed, or measured, and one assessed by inference, or computation.

## *Observables*

We formulate all variables that we assess as random variables. For example, the acoustic signal at its arrival at the human ear is a random variable, that of the varying pressure of air when it is compressed or rarefied.

We may analyze this random variable in various ways. The result may be a multitude of random variables. In our example of the acoustic signal, we may perform some sort of frequency analysis. The result would be a 2-dimensional array of random variables, one at each chosen frequency and instant of time over a specified period of time. Such a collection of random variables, all taking on values from the same value space, is called a random field. The example of the varying air pressure is a degenerate case where the field is a singleton. We can subject the first random field to alternative analysis, or we can subject the derived random field to further analysis, to generate other random fields. We call these random fields observables.

We define a system of observation as a finite set of such observables. We define perception as a special subset of such systems of observation in which there are *target* observables. A target observable is a 1-dimensional random field, where the value space is a finite discrete set. It is a target in the sense that the objective of the perception is the determination of this random field. We will sometimes call them targets, for short. The one dimension is usually time. A simple example might be a random variable of two elements, one is represented by the phrase "there is a cheetah", and one by the phrase "there is no cheetah". A more complex example is a spoken language, where the target random variable is the repertory of gestures of articulation.

In principle, we can ascertain the probability distribution of each observable, except the target, by measurements and computations. Perception as a problem for scientific inquiry is solving the as-yet unknown computations for inferring the target observable. In any experiment, or ensembles of repeated observations, we mandate the distribution of the target in accordance with the design of the experiment, or it is dictated by the circumstances requiring the observations. For speech perception, we do so by the choice of the articulatory gestures performed, or the playback of recordings of such performances.

## *Mutual Information*

Let the random field A represent the target, and Ω be one of the observables derivable from some measurement related to the target. Each set of values of a random field is called a configuration. We will assume that the random variables of the random fields A and Ω may be indexed by a finite set of integers respectively[3]. The notation is that by default, we use the upper case font to represent a set, and the lower case font a corresponding member of the set. We will explicitly declare a symbol when we need to disambiguate.

---

3   This does not restrict the dimensionality of the domain of the random fields.

A configuration of Ω is the set of instances of the random variables:

$$\{\ldots, \omega_1, \omega_2, \ldots, \omega_k, \ldots\}$$

where the subscript k stands for a member of some indexing set of integers, K. For notational parsimony, we will use the more convenient $\omega_K$ as equivalent. Likewise, for the configuration space of A, $\alpha_J$. We will denote the corresponding configuration space, which consists of the product space of the value spaces of the random variable at every index in the domain of the random field as $\Omega_K$ and $A_K$.

We regard a random variable as "continuous" if all probability distributions and arithmetic operations remain well defined for any sampling resolution of interest. For perception, there is always a finite upper limit to the sampling resolution.

The probability distribution of a random field is over the configuration space into the real-valued interval [0,1]:

$$\Omega_K \to [0,1] : p(\omega_K) \quad,$$

$$A_J \to [0,1] : p(\alpha_J) \quad.$$

Use of the same letter 'p' is again merely notational convenience, and does not imply that the distributions are the same function.

We regard a random variable as "continuous" if all probability distributions and arithmetic operations remain well defined for any sampling resolution of interest. For perception, there is a finite upper limit to the sampling resolution.

The random field of the joint observation of the random fields A and Ω is denoted AΩ. The configuration space of AΩ is just the product space of the configuration spaces of A and Ω:

$$\alpha_J \omega_K = \{\ldots, \alpha_1, \alpha_2, \ldots, \alpha_j, \ldots, \omega_1, \omega_2, \ldots, \omega_k, \ldots\} \quad.$$

In the context of the joint random field, $p(\omega_K)$ and $p(\alpha_J)$ now denote their marginal distributions:

$$p(\omega_K) = \sum_{\alpha_J \in A_J} p(\alpha_J \omega_K) \quad,$$

where the summation is over the configuration space $A_J$; and likewise :

$$p(\alpha_J) = \sum_{\omega_K \in \Omega_K} p(\alpha_J \omega_K) \quad.$$

We may now write the entropy of A and Ω as

$$E(A) = \sum_{\alpha_J \in A_J} p(\alpha_J) \ln p(\alpha_J) \quad,$$

$$E(\Omega) = \sum_{\omega_K \in \Omega_K} p(\omega_K) \ln p(\omega_K) \quad.$$

For the joint random field AΩ,

$$E(A\,\Omega) = \sum_{\alpha_J \omega_K \in A\,\Omega} p(\alpha_J \omega_K) \ln p(\alpha_J \omega_K) \quad . \tag{1}$$

We will interpret the entropy as quantifying the amount of variability in the configurations of a random field. That the entropy is numerically less than or equal to 0 is regarded as a convention, meaning that if it is re-defined to be positive definite, all statements retain their truth values. But interpretively, its numerically negative definiteness suggests the state of indeterminacy prior to any observation when we have a *deficit* of information. When we reach a state of certainty about A, our deficit of information about A is canceled out as a result of gaining an amount of information $I(A)$, such that

$$E(A) + I(A) = 0 \quad . \tag{2}$$

The amount of variability in the jointly observed AΩ together may be *less* than the sum of the variability in A or Ω separately, if certain configurations of Ω tend to occur together with certain configurations of A; in other words, if A and Ω are correlated.

The problem of perception becomes: what can we infer about the configurations the random field A may take on, when we have observed that the random field Ω has the configuration $\omega_K$? Let us first denote the subspace of $A_J$ given $\omega_K$

$$A_J | \omega_K \quad .$$

The distribution of A given a particular $\omega_K$ is called the conditional distribution of A with respect to $\omega_K$:

$$p(\alpha_J | \omega_K) = \frac{p(\alpha_J \omega_K)}{p(\omega_K)} \quad . \tag{3}$$

So we can define the remaining deficit in information about A, or the residual entropy of A given $\omega_K$ in terms of this conditional distribution:

$$E(A | \omega_K) = \sum_{\alpha_J \in A_J | \omega_K} p(\alpha_J | \omega_K) \ln p(\alpha_J | \omega_K) \quad . \tag{4}$$

By observing, $\omega_K$ we have not reached certainty about A, but we have gained an amount of information, $I(A | \omega_K)$, such that the remaining deficit in information about A is $E(A | \omega_K)$:

$$E(A) + I(A | \omega_K) = E(A | \omega_K) \quad . \tag{5}$$

In principle, we can arrange a series of experiments, in which we record the instances of the joint occurrence of $\alpha_J \omega_K$ for all possible configurations of Ω, assuming that we can either directly perform $\alpha_J$, or be informed by some oracle (human). The expected gain in information, which we denote by $I(A | \Omega)$, over such a series of experiments would be

$$I(A | \Omega) = \sum_{\omega_K \in \Omega_K} p(\omega_K) I(A | \omega_K) \quad . \tag{6}$$

From (4) and (5),
$$I(A|\Omega)=E(A|\Omega)-E(A) \quad,\tag{7}$$
where
$$E(A|\Omega)=\sum_{\omega_K \in \Omega_K} p(\omega_K) E(A|\omega_K) \quad.\tag{8}$$
is the expected residual entropy about A when we observe Ω. Further, from (1), (2) and (3),
$$I(A|\Omega)=E(A\,\Omega)-E(\Omega)-E(A) \quad.$$

The expected gain in information of A by observing Ω was termed the correlation information [Everett1], but is now called mutual information[Wikipedia1]. The non-negativity of the mutual information can be proven in general[Everett1].

If there is *no* tendency for certain configurations of A to occur with certain configurations of Ω, or A is independent of Ω, then we should expect to gain zero information, a well known result.

Otherwise, if A is correlated with Ω to any degree, we would expect to gain some information. Furthermore, the mutual information may not exceed the lesser of the information that can be gained about A or Ω separately. This is because the entropy of a joint random field is always less than or equal to the entropy of each random field separately[4]
$$E(A\,\Omega)\leq E(\Omega) \quad,\quad E(A\,\Omega)\leq E(A) \quad.$$

For perception, the target A, as a finite discrete random field, has less variability than Ω, and therefore
$$I(A|\Omega)\leq -E(A) \quad.$$

In other words, the maximum amount of information we can expect to gain about A from Ω cannot cancel out more than the entropy of A. In the case of maximal $I(A|\Omega)$,
$$I(A|\Omega)=-E(A) \quad.$$

By (1), the maximum mutual information is equal to the information that can ever be gained about A,
$$I(A|\Omega)=I(A) \quad.$$

And by (7), the expected residual entropy of A is 0
$$E(A|\Omega)=0 \quad.\tag{9}$$

---

4 Regard the joint random field as a fine-graining of its separate random fields. The entropy monotonically decreases under fine-graining [Everett1].

## *Maximally Informative Observables*

We will now show that (9) is true if and only if the conditional distribution $p(\alpha_J|\omega_K)$ assumes the values 0 or 1. Expanding (9) with (4),

$$E(A|\Omega) = \sum_{\omega_K \in \Omega_K} \sum_{\alpha_J \in A|\omega_K} p(\omega_K) p(\alpha_J|\omega_K) \ln p(\alpha_J|\omega_K) \quad . \tag{10}$$

Each term in this sum is $\leq 0$. Therefore the sum = 0 if and only if each term = 0. This means that except for a subset of measure 0 ($p(\omega_K) = 0$), either $p(\alpha_J|\omega_K) = 0$, or $\ln p(\alpha_J|\omega_K) = 0$, in which case $p(\alpha_J|\omega_K) = 1$.

This means upon observing any configuration of Ω, we can conclude that there is one and only one particular configuration A can assume. We say the random field Ω is *maximally informative* about the random field A, or Ω is a *maximally informative observable(MIO)* and the target A is maximally observable with respect to Ω.

In addition, let us postulate that the configuration space $\Omega_K$ allows us to define neighborhoods in it and that the conditional distribution $p(\alpha_J|\omega_K)$ is a piecewise continuous function of $\Omega_K$, all reasonable assumptions for the problem of perception. Then for each configuration $\alpha_J$, there exists in the configuration space $\Omega_K$ a neighborhood $\nu(o_K|\alpha_J)$ of some $o_K$ [5] such that

$$p(\alpha_J|\omega_K) = \begin{cases} 1, \forall \omega_K \in \nu(o_K|\alpha_J) \\ 0, \text{otherwise} \end{cases} \quad . \tag{11}$$

This is the experimentally observed identification function of categorical perception.

## *Model Distributions*

The most important property of a maximally informative observable is that the indicator form of $p(\alpha_J|\omega_K)$ is independent of the marginal distribution of A, or of Ω. By definition (3), the conditional distribution of Ω given A is given by

$$p(\omega_K|\alpha_J) = \begin{cases} \dfrac{p(\omega_K)}{p(\alpha_J)}, \forall \omega_K \in \nu(o_K|\alpha_J) \\ 0, \text{otherwise} \end{cases} \quad . \tag{12}$$

Interpretively, these are the "models" of the configurations of A in terms of what configurations of Ω each configuration of A tends to give rise to. We find that the support of the distribution of Ω for each configuration $\alpha_J$ is restricted to $\nu(o|\alpha_J)$, is distinct from that of any other configuration and that the distribution is proportional to the marginal distribution $p(\omega_K)$, the factor of proportionality being the inverse of the marginal probability $p(\alpha_J)$. Regardless of the target observable's distribution, the model distributions are of the same form. Summing over both sides of (12):

$$\sum_{\omega_K \in \nu(o|\alpha_J)} p(\omega_K) = p(\alpha_J) \quad . \tag{13}$$

---

5   In general, there could be more than one such neighborhood for each configuration of A, not necessarily connected. All statements we make here about the one neighborhood apply to all such neighborhoods without qualifications.

Thus varying the target distribution serves only to vary the relative occurrences of the configurations in the distinct regions of support by the same proportions. This makes the physical system comprising Ω and A eminently suitable for use in communication. For speech, the distribution of A may be arbitrary per ensemble of speech gestures, such as a particular language, or even a particular practical application.

Further, (13) implies that the conditional distribution $p(\alpha_J|\omega_K)$ has induced a partition of the configuration space $\Omega_K$ by the configuration space $A_J$ such that the probability for the occurrence of any configuration in each partition is equal to the probability of the corresponding configuration of A, thus A is a coarse-graining of Ω.

## *Sub-Maximally Informative Observables*

It is easy to see that a somewhat less "perfect" coarse-graining of Ω can be achieved with desirable properties similar to (11), (12) and (13), that would reduce the residual entropy $E(A|\Omega)$ substantially but short of zero.

Let $\{B_J^l | l \in L\}$ be a partition of $A_J$. Then the conditional distribution

$$p(B_J^l|\omega_K) \equiv \sum_{\alpha_J \in B_J^l} p(\alpha_J|\omega_K) = \begin{cases} 1, \forall \omega_K \in \nu(o|\alpha_J \in B_J^l) \\ 0, \text{ otherwise} \end{cases}$$

would induce a partition of $\Omega_K$ such that

$$p(\omega_K|B_J^l) \equiv \sum_{\alpha_J \in B_J^l} p(\omega_K|\alpha_J) p(\alpha_J) = \begin{cases} p(\omega_K), \forall \omega_K \in \nu(o|\alpha_J \in B_J^l) \\ 0, \text{ otherwise} \end{cases}$$

$$\sum_{\omega_K \in \nu(o|\alpha_J \in B_J^l)} p(\omega_K) = p(B_J^l)$$

We call an observable with such a conditional distribution a sub-maximally informative observable (sub-MIO). It induces a coarser coarse-graining than a MIO. Several such sub-MIOs, however, can jointly provide a finer coarse-graining equivalent to, or even redundant to that of an MIO.

## *Implications for Speech Perception*

The original (and subsequent variations of) categorical perception of plosive consonants is a very constrained experimental design in which the target is a single random variable, the place of articulation of 3 plosive consonants, and the putative MIO is also a single random variable, the change in the frequency of the second formant. All other degrees of freedom are frozen. The full problem of speech perception is more complex. The change in frequency can be regarded as the projection of the neighborhood, $\nu(o|\alpha)_{K\ J}$, which is a multidimensional space, onto a one-dimensional subspace. The full neighborhood could be a rather complicated blob in configuration space. The categorical property of the identification function could become attenuated as more degrees of freedom are involved. This does not nullify the original observed categorical perception, but it is likely that the selected observable, the change in the frequency of the second formant, is not the MIO, or a sub-MIO of speech gestures, but its shadow.

In general, we would expect a complex target like speech gestures to be sub-maximally observable via a multitude of sub-MIOs. Such redundancy would enable the identification of a target from noise. This may be the reason that the neurophysiology of audition is now known to be replete with multidimensional random fields from the cochlea up to and including the primary auditory cortex[Wang1]. Note that the sub-MIOs need not be orthogonal, or even independent. The experimental results on categorical perception of speech imply that there is at least one sub-MIO.

Given a physical system in nature, it is highly unlikely that all of its observables are maximally informative. In fact, it is unlikely that any of its observables is maximally informative. However, even a sub-maximally informative observable is useful if the target to be inferred, such as the presence or absence of a cheetah in the tall grass on the savannah, is incidental, the penalty for failure to infer possibly fatal, and the cost of false alarms not very high.

If speech has evolved for human to human communication, then there would have been selective pressure for the co-evolution of the vocal tract with the auditory neurophysiology to craft observables that tend asymptotically to be maximally informative on the target.